\documentclass[10pt,twocolumn,letterpaper]{article}

\usepackage[pagenumbers]{iccv} 

\definecolor{iccvblue}{rgb}{0.21,0.49,0.74}
\usepackage[pagebackref,breaklinks,colorlinks,allcolors=iccvblue]{hyperref}
\usepackage{algorithm}
\usepackage{algorithmic}
\usepackage{booktabs}
\usepackage{multirow}
\usepackage{threeparttable}
\usepackage{makecell}
\usepackage{amsmath}
\usepackage{amsfonts}
\usepackage{pifont}
\usepackage{color}
\usepackage{colortbl}
\usepackage{newfloat}
\usepackage{listings}
\floatstyle{ruled}
\newfloat{listing}{tb}{lst}{}
\floatname{listing}{Listing}
\def\paperID{*****} 
\def\confName{ICCV}
\def\confYear{2025}

\title{Temporal Reversal Regularization for Spiking Neural Networks:\\
Hybrid Spatio-Temporal Invariance for Generalization}

\author{
	Lin Zuo,
    Yongqi Ding,
    Wenwei Luo,
    Mengmeng Jing,
    Kunshan Yang\\
    University of Electronic Science and Technology of China\\
{\tt\small linzuo@uestc.edu.cn,yqding@std.uestc.edu.cn,wenweiluo2022@163.com}\\
{\tt\small jingmeng1992@gmail.com,ksyang@std.uestc.edu.cn}
}

\begin{document}
\maketitle
\begin{abstract}
Spiking neural networks (SNNs) have received widespread attention as an ultra-low power computing paradigm. Recent studies have shown that SNNs suffer from severe overfitting, which limits their generalization performance. In this paper, we propose a simple yet effective Temporal Reversal Regularization (TRR) to mitigate overfitting during training and facilitate generalization of SNNs. We exploit the inherent temporal properties of SNNs to perform input/feature temporal reversal perturbations, prompting the SNN to produce original-reversed consistent outputs and learn perturbation-invariant representations. To further enhance generalization, we utilize the lightweight ``star operation” (Hadamard product) for feature hybridization of original and temporally reversed spike firing rates, which expands the implicit dimensionality and acts as a spatio-temporal regularizer. We show theoretically that our method is able to tighten the upper bound of the generalization error, and extensive experiments on static/neuromorphic recognition as well as 3D point cloud classification tasks demonstrate its effectiveness, versatility, and adversarial robustness. In particular, our regularization significantly improves the recognition accuracy of low-latency SNN for neuromorphic objects, contributing to the real-world deployment of neuromorphic computational software-hardware integration.
\end{abstract}

\section{Introduction}
\label{sec:intro}

Recently, brain-inspired spiking neural networks (SNNs) have received widespread attention. Unlike traditional artificial neural networks (ANNs), which transfer information using intensive floating-point values, SNNs transfer discrete 0-1 spikes between neurons, providing a more efficient neuromorphic computing paradigm~\cite{Spike_driven_Transformer}. In addition, spiking neurons, which simulate the dynamics of biological neurons over multiple timesteps, can effectively extract temporal features~\cite{kim2023exploring}. These superior properties have led to the application of SNNs to a variety of spatio-temporal tasks such as object recognition, detection, generation, and natural language processing~\cite{Su_2023_ICCV,kamata2022fully,bal2024spikingbert}.

To improve the performance of SNNs, researchers have made considerable efforts to enhance their feature extraction ability. For instance, the temporal properties of SNNs are optimized through heterogeneous timescales~\cite{chakraborty2024sparse}, batch normalization (BN) methods adapted to the temporal dimension~\cite{TEBN,TAB}, and improved neuron dynamics~\cite{ponghiran2022spiking}. Alternatively, the spatial properties of SNNs are continuously improved with sophisticated network architectures~\cite{Spike_driven_Transformer,QKFormer}. 

However, recent studies have shown that SNNs suffer from severe overfitting. For instance, the gap between training and test accuracy of the SOTA SNN~\cite{Spike_driven_Transformer,QKFormer} on the neuromorphic dataset CIFAR10-DVS is almost 20\%. Even with carefully designed data augmentation strategies~\cite{NDA,EventAugment,EventDrop}, overfitting remains severe, especially for low latency (timestep) SNNs~\cite{MPS}. This suggests that there is an urgent need in the community for more generalizable strategies to mitigate the overfitting of SNNs.

In this paper, we propose a simple yet effective temporal reversal regularization (TRR) to mitigate the overfitting of SNNs and improve the generalization performance.  We perturbed the SNN during training, pushing it to be robust to these perturbations and to focus on generalizable features. Specifically, we explore the inherent temporal properties of SNNs, perform temporal reversal perturbation of input/spike features for temporal/static tasks, and generate the corresponding pair of outputs during training. We encourage this pair of outputs to be as similar as possible, allowing the SNN to learn time-invariant generalized spatial representations on the one hand, and perturbation-insensitive stable temporal representations on the other hand. 

To further facilitate generalization, we employ the lightweight ``star operation" (Hadamard product) to hybridize the high-dimensional original and temporally reversed features, expanding the implicit dimensions and prompting the SNN to make correct predictions for the hybrid features. The feature hybridization further perturbs the temporal dimension and disrupts the spatial feature map, which can be considered as a regularization of high-dimensional features (visualization in \textbf{Appendix D in the Supplementary Material}), allowing the SNN to learn latent representations that are insensitive to spatio-temporal perturbations.

We theoretically demonstrate that the proposed method can tighten the upper bound of the generalization error and thus improve the generalization performance through the PAC-Bayesian framework~\cite{10.1007/978-3-540-45167-9_16,NIPS2017_10ce03a1}. To confirm the effectiveness of our method, we conducted extensive experiments using VGG, ResNet, Transformer, and PointNet architectures on static object recognition, neuromorphic object/action recognition, and 3D point cloud classification tasks. The experimental results show that our method achieves consistent performance gains with excellent generalizability, and also improves the adversarial robustness of SNNs. In summary, our contributions are as follows:
\begin{itemize}
    \item We propose to exploit the temporal properties of SNNs for temporal reversal to generate original-reversed consistent outputs to mitigate overfitting and improve generalization.
    \item We propose to hybridize the high-dimensional original and reversed spike firing rates by a lightweight ``star operation" as a spatio-temporal regularization to further facilitate the generalization of the SNN.
    \item We demonstrate the ability of the proposed method to tighten the upper bound of the generalization error as well as its effectiveness, generality, and robustness gain through theoretical analysis and extensive experiments.
\end{itemize}

\section{Related Work}

\textbf{Spiking Neural Network.} To train high-performance SNNs, indirect training based on ANN-to-SNN conversion~\cite{9543525,hao2023bridging} and direct training method based on surrogate gradient~\cite{STBP,guo2024ternary,qiu2024gated} have achieved remarkable results. In addition, improved BN strategies~\cite{TEBN,TAB}, neuron dynamics~\cite{ponghiran2022spiking,PALIF,wang2024autaptic}, and sophisticated architectures borrowed from ANNs~\cite{Spike_driven_Transformer,10535518} further boost the performance of SNNs. Despite recent progress, SNN still suffers from severe overfitting~\cite{Spike_driven_Transformer,QKFormer,MPS}, and even the incorporation of data augmentation strategies~\cite{NDA,EventDrop,EventAugment} is not sufficient to alleviate this problem. To this end, we introduce perturbations into SNNs for regularization in a simple yet effective way to mitigate overfitting. In addition, our regularization is architecture- and task-independent with superior versatility.

\textbf{Self-Supervised Learning.} Our method shares a similar philosophy to self-supervised learning (SSL). For a given input, SSL uses data augmentation to transform it into two different views and increase the similarity between the outputs generated by the two~\cite{Chen_2021_CVPR,Wang_2022_CVPR}. This can facilitate the neural network to learn generalized representations that are invariant to data transformations~\cite{pmlr-v119-lee20c,pmlr-v119-chen20j}. However, SSL is extremely sensitive to data augmentation strategies~\cite{pmlr-v119-chen20j}. For our method, we avoid the tedious process of data augmentation search and use the inherent temporal property of SNNs for perturbation to improve model performance. From another perspective, our method can be seen as an extension of SSL in SNNs to learn generalized spatio-temporal representations.

\textbf{Data Augmentation in SNNs.} Data augmentation can slightly alleviate overfitting. For static object recognition, SNNs typically use augmentation strategies such as random flipping and cropping, which are adopted from ANNs~\cite{Spike_driven_Transformer,qiu2024gated}. For neuromorphic data with temporal properties, these augmentation strategies are not applicable and require special design~\cite{NDA,EventDrop,EventAugment}. This makes it cumbersome for existing SNNs to adapt the augmentation strategy to the task at hand. Moreover, even with data augmentation, SNNs still suffer from severe overfitting~\cite{Spike_driven_Transformer,QKFormer}. Instead, our temporal reversal can be viewed as a generalized augmentation strategy that exploits the temporal properties of the SNN. In addition, temporal reversal outperforms existing data augmentation methods and can be further extended in conjunction with them.

\begin{figure*}[t]
\centering
\includegraphics[width=0.9\textwidth]{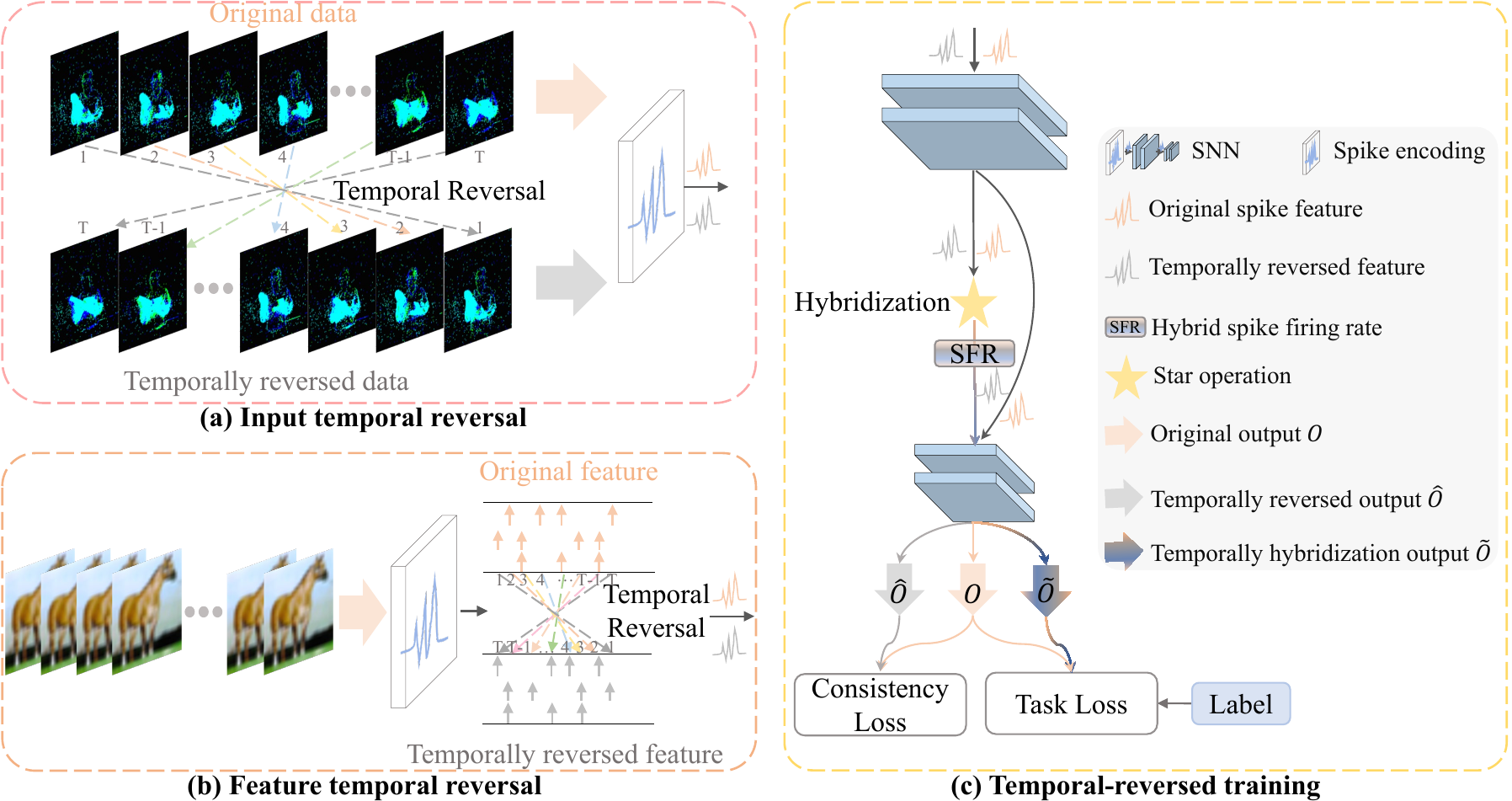}
\vskip -0.1in
\caption{Overview of the proposed method. TRR perturbs temporal and static data by (a) input and (b) spike feature temporal reversal, allowing the SNN to produce original and temporally reversed outputs, and (c) encouraging both outputs to be as similar as possible to learn generalized perturbation-invariant representations. In addition, TRR hybridizes the original and temporally reversed spike firing rates and expands the implicit dimensionality with a lightweight ``star operation”. This serves as a spatio-temporal regularizer that further facilitates the generalization of SNNs.}
\label{overview}
\vskip -0.25in
\end{figure*}

\section{Method}

\subsection{Spiking Neuron Model}
Spiking neurons iteratively receive input currents, accumulate in membrane potentials, and generate spikes. For the most commonly used leaky integrate-and-fire (LIF) model~\cite{STBP}, the dynamics of the accumulating membrane potential can be expressed as:
\vskip -0.1in
\begin{equation}
H_{i}^{l}(t)=\left(1-\frac{1}{\tau}\right) H_{i}^{l}(t-1)+I_{i}^{l}(t),
\end{equation}
where $H$ and $I$ denote the membrane potential and afferent current, respectively, $l$ and $i$ are the layer and neuron index, $t$ is the timestep, and $\tau$ is the time constant controlling the degree of leakage of membrane potential.

When the membrane potential $H_{i}^{l}(t)$ reaches the firing threshold $\vartheta$, the spiking neuron will generate a spike $S_{i}^{l}(t)$ and reset the membrane potential:
\vskip -0.1in
\begin{equation}
S_{i}^{l}(t) = \left\{
\begin{array}{cl}
1,\quad H_{i}^{l}(t) \ge \vartheta \\
0,\quad H_{i}^{l}(t) < \vartheta \\
\end{array},
\right.
\label{eqfire}
\end{equation}
\vskip -0.1in
\begin{equation}
H_{i}^{l}(t) = H_{i}^{l}(t)-S_{i}^{l}(t)\vartheta.
\end{equation}
\vskip -0.05in
Since the spike activity is non-differentiable, we replace the spike derivatives with the surrogate gradient during backpropagation to optimize the SNN using the Backpropagation Through Time (BPTT) algorithm. In this paper, we use the rectangular surrogate function:
\vskip -0.2in
\begin{equation}
\frac{\partial S_{i}^{l}(t)}{\partial H_{i}^{l}(t)} \approx \frac{\partial h(H_{i}^{l}(t), \vartheta)}{\partial H_{i}^{l}(t)} = \frac{1}{a} \text{sign} (|H_{i}^{l}(t)-\vartheta|<\frac{a}{2}),
\end{equation}
where $a$ is set to 1.0 and is a hyperparameter that controls the shape of the surrogate function.

\subsection{Temporal Reversal Perturbation}

In this section, we present how temporal reversal perturbation mitigates overfitting and promotes the generalizability of SNNs.

\textbf{Input Temporal Reversal.} Formally, we denote the input data with temporal properties as $X=\{x_1,x_2,\cdots,x_T\} \in \mathbb{R}^{T \times B \times C_{in} \times H_{in} \times W_{in}}$, where $T,B,C_{in},H_{in}$, and $W_{in}$ are the time, batch, input channel, height and width sizes, respectively. Typically, the temporal input $X$ and the temporal dimension of the SNN are aligned, i.e., $x_t$ is input to the SNN at timestep $t$ and ultimately produces the output $o_t$. In addition to the original input $X$, we use temporal reversal to additionally generate the temporally reversed input $\hat{X}\!\!\!=\!\!\! \{\hat{x}_1,\!\hat{x}_{2},\!\cdots\!,\!\hat{x}_T\}$ for perturbation. As shown in Fig.~\ref{overview}~(a), this temporal reversal is achieved by simply flipping the temporal index of the input data X, i.e., $\hat{x}_{t}=x_{T+1-t}$, without laboriously selecting data augmentations to generate additional data views as in self-supervised learning~\cite{Chen_2021_CVPR,Wang_2022_CVPR}. At each timestep $t$, $\hat{x}_t$ is fed into the SNN to produce the temporally reversed output $\hat{o}_t$.

\textbf{Feature Temporal Reversal.} Input temporal reversal can only be used for tasks with inherent temporal properties, such as neuromorphic or video data. To make this temporal reversal to be effective for static tasks without inherent temporal properties, we further propose feature temporal reversal. For static data $x \in \mathbb{R}^{B \times C_{in} \times H_{in} \times W_{in}}$, SNNs typically input data repeatedly at each timestep and encode it as spikes through the first spiking neuron layer. We denote the primary features after spike encoding by $F=\{f_1,f_2,\cdots,f_T\} \in \mathbb{R}^{T \times B \times C_F \times H_F \times W_F}$, where $f_t$ represents the encoded spikes at timestep $t$. With this, we take advantage of the spiking neuron dynamics to transform the static data $x$ into spatio-temporal spikes $F$ with the temporal dimension. We then apply temporal reversal to the spike feature $F$ and obtain the temporally reversed feature $\hat{F}=\{\hat{f}_1,\hat{f}_{2},\cdots,\hat{f}_T\}$, where $\hat{f}_t=F_{T+1-t}$, as shown in Fig.~\ref{overview} (b). The temporal reversed feature $\hat{F}$ is propagated further forward in the SNN to produce the final temporally reversed output $\hat{O}=\{\hat{o}_1,\hat{o}_2,\cdots,\hat{o}_T\}$.

\textbf{Perturbation-Invariant Learning.} Through input/feature temporal reversal, we can perturb the temporal dimension of the SNN, regardless of whether the input is inherently temporal or not. To motivate the SNN to learn perturbation-invariant spatio-temporal features, we impel the temporal-reversed output $\hat{O}$ to be as similar as possible to the original output $O$. As shown in Fig.~\ref{overview} (c), we increase the similarity between the two by imposing a consistency loss $\mathcal{L}_{con}$.

We illustrate the consistency loss in detail with a $C$-way classification task. For the outputs $O$ and $\hat{O}$ of the SNN, the corresponding category probabilities are given as:
\vskip -0.22in
\begin{equation}
p_j=\frac{e^{z_j/T_{tem}}}{\sum^C_{c=1}{e^{z_c/T_{tem}}}},\hat{p}_j=\frac{e^{\hat{z}_j/T_{tem}}}{\sum^C_{c=1}{e^{\hat{z}_c/T_{tem}}}},
\end{equation}
where $z=\frac{1}{T}\sum^{T}_{t=1}o_t$, $\hat{z}=\frac{1}{T}\sum^{T}_{t=1}\hat{o}_t$ are the rate-decoded output logits and the subscript $j$ denotes the $j$-th class. $T_{tem}$ is the temperature scaling hyperparameter used to smooth the logit, which is set to 2 in this paper. We use KL divergence to push the category probabilities of the reversed output to be consistent with the probability distribution of the original output:
\vskip -0.22in
\begin{equation}
\mathcal{L}_{con}=T_{tem}^2KL(p||\hat{p})=T_{tem}^2\sum^{C}_{j=1}{p_jlog(\frac{p_j}{\hat{p}_j})}.
\end{equation}
\vskip -0.1in
Thus, as the SNN is trained, both task loss (cross-entropy loss $\mathcal{L}_{CE}$) and consistency loss $\mathcal{L}_{con}$ contribute to the optimization of the parameters:
\begin{equation}
\hat{\mathcal{L}}=\mathcal{L}_{CE}(O,Y) + \mathcal{L}_{con}(O,\hat{O}),
\label{TRloss}
\end{equation}
where $Y$ is the ground-truth label.

It is worth noting that although static data is not temporally featured, the inherent temporal dynamics of the SNN makes the spike sequence temporally featured, so that temporal reversal still contributes to the static task. While for time-sensitive actions such as left-to-right hand waving, the reversed data causes temporal confusion, Eq.~\ref{TRloss} motivates the SNN to learn the correct temporal order based only on the original output and the stable spatial representation based on the synergistic learning of the two. The ablation studies in Tab.~\ref{ablation:vgg} confirm the effectiveness of temporal reversal in various tasks.

\subsection{Feature Hybridization Perturbation}
To further improve the generalization of the SNN, we propose to hybridize original and temporally reversed features for perturbation. The lightweight ``star operation" (Hadamard product) can significantly increase the implicit dimensionality of ANN features, and shares a philosophy with kernel functions~\cite{Ma_2024_CVPR,shawe2004kernel}. Therefore, we propose to perturb the original and temporal reversed features in the SNN with the ``star operation" to serve as a spatio-temporal regularization of the high-dimensional features. However, due to the binary nature of the spike, the ``star operation" does not contribute to dimensionality expansion in SNNs, but instead causes severe information loss. In the following, we will analyze this problem and make the ``star operation" in SNNs feasible by converting spikes into firing rate. 

\textbf{Information Loss in SNNs with Star Operation.} For brevity, similar to~\cite{Ma_2024_CVPR}, we write the ``star operation" as $(W^T_1X+B_1) \ast (W^T_2X+B_2)$. Representing $X=\begin{bmatrix} X\\1\end{bmatrix}$, $W=\begin{bmatrix} W\\B\end{bmatrix}$ in matrix form, the star operation becomes $(W^T_1X)\ast (W^T_2X)$. We focus on the ANN scenario with one output channel and a single-element input, i.e., consider $w_1$, $w_2$, and $x \in~\mathbb{R}^{(d+1) \times 1}$, where $d$ denotes the input channel number (which can be naturally extended to scenarios with multiple output channels and multiple input elements). The ``star operation" can be rewritten as:
\vspace{-0.1in}
\begin{equation}
\begin{aligned}
& w_{1}^{\mathrm{T}} x * w_{2}^{\mathrm{T}} x 
=  \left(\sum_{i=1}^{d+1} w_{1}^{i} x^{i}\right) *\left(\sum_{j=1}^{d+1} w_{2}^{j} x^{j}\right) \\
= & \underbrace{\alpha_{(1,1)} x^{1} x^{1} \! \! + \! \cdots \! + \! \!\alpha_{(4,5)} x^{4} x^{5} \! \! + \! \cdots \! + \! \!\alpha_{(d+1, d+1)} x^{d+1} x^{d+1}}_{\textcolor[RGB]{161, 163, 166}{(d+2)(d+1) / 2 \text { items }}}
\end{aligned}
\end{equation}
where $i$, $j$ are the channel indices and $\alpha$ is the coefficient of each element:
\vspace{-0.1in}
\begin{equation}
\alpha_{(i,j)}=\left\{\begin{array}{cc}
w_{1}^{i} w_{2}^{j} & \text { if } i=j \\
w_{1}^{i} w_{2}^{j}+ w_{1}^{j} w_{2}^{i} & \text { else }
\end{array}\right..
\end{equation}
As a result, the “star operation” in ANNs is able to transform the $d$-dimensional feature $x$ into $\frac{(d+2)(d+1)}{2}$ distinct elements, each of which, except $\alpha_{d+1,:}x^{d+1}x$, is nonlinearly associated with $x$, serving as a dimensionality expansion~\cite{Ma_2024_CVPR}.

However, unlike ANNs, there are negative consequences of directly using the ``star operation" hybridization in SNNs. Since spiking neurons generate binary spikes, the features $X \in \mathbb{B}$, where $\mathbb{B}$ is the binary set. Thus, dimensional expansion and nonlinear combination for $x$ results in $x^ix^j \in \mathbb{B}$ where $i,j=\{1,2,\cdots,d+1\}$. This means that $x^ix^j$ can only take values in the binary ${0,1}$, and that the ``star operation" does not work for dimensional expansion. In addition, due to the inherent sparsity of spikes, most of the features in the SNN are 0, with very few 1-valued spikes. The Hadamard product between binary spikes results in increased 0-value spikes, since 1 is only output if both sides are 1:
\vskip -0.2in
\begin{equation}
x^ix^j=\left\{\begin{array}{cc}
1 & \text { only if } x^i=x^j=1 \\
0 & \text { else }
\end{array}\right..
\end{equation}
This makes the spikes even sparser and reduces the expressiveness of the SNN, leading to performance degradation.

\textbf{Star Operation on Spike Firing Rate.} To avoid performance degradation caused by ``star operations" on 0-1 spikes, we convert multiple timestep spikes $\{x_1,x_2,\cdots,x_T\} \in \{0,1\}$ to spike firing rate $\Phi=~\frac{1}{T}\sum^{T}_{t=1}x_t$. The spike firing rate is spaced at $\frac{1}{T}$ intervals and takes on the value range $[0,1]$, which can be viewed as a multi-bit value, greatly improving its representability compared to binary spikes. For instance, $\Phi$ can be taken as $\{0,\frac{1}{5},\frac{2}{5},\frac{3}{5},\frac{4}{5},1\}$ at $T=5$. In this way, employing the ``star operation" on the spike firing rate can take advantage of the dimensional expansion benefits it is supposed to provide and avoid the degradation of the SNN due to excessive 0-value outputs.

In practice, we use the ``star operation" to hybridize the original and temporally reversed spike firing rates of the penultimate layer of the SNN, which is passed directly to the final classification layer, as shown in Fig.~\ref{overview} (c). In this way, the SNN produces two outputs: the original output $O$ with the temporal dimension and the temporally hybridization output $\tilde{O}$ without the concept of time. We guide both outputs with label $Y$ to facilitate the SNN to ignore ``star" perturbations due to hybridization and learn more generalized representations:
\vskip -0.15in
\begin{equation}
\tilde{\mathcal{L}}=(1-\alpha)\mathcal{L}_{CE}(O,Y) + \alpha\mathcal{L}_{CE}(\tilde{O},Y),
\label{FHloss}
\end{equation}
where $\alpha$ is the balance coefficient, which will be analyzed in the experimental section.

\begin{algorithm}[tb]
\caption{Temporal reversal regularization for SNNs}
\label{alg:trt_training}
\textbf{Input}: input data $x$, label $Y$.\\
\textbf{Parameter}: timestep $T$, balance coefficient $\alpha$.\\
\textbf{Output}: Trained $n$-layer SNN.
\begin{algorithmic}[1] 
\STATE Initialize SNN parameters $\theta$
\FOR{$i=1,2,\cdots,I_{train}$ iteration}
\IF {$x$ with time dimension}
\STATE $\hat{x} = f_{re}(x)$ ;       // Input temporal reversal\\
\STATE $F=E(x),\hat{F}=E(\hat{x})$;       // Spike encoding\\
\ELSE
\STATE $F=E(x)$;       // Spike encoding\\
\STATE $\hat{F} = f_{re}(F)$ ;       // Feature temporal reversal\\
\ENDIF
\STATE $F^{n-1}=SNN(F),\hat{F}^{n-1}=SNN(\hat{F})$;       // Forward propagation of features to the penultimate layer\\
\STATE $\tilde{\Phi}=\Phi^{n-1} \ast \hat{\Phi}^{n-1}=\frac{1}{T}\sum^{T}_{t=1}F^{n-1}_t \ast \frac{1}{T}\sum^{T}_{t=1}\hat{F}^{n-1}_t$;       // Spike firing rate hybridization\\
\STATE $O=fc(F^{n-1}),\hat{O}=fc(\hat{F}^{n-1}),\tilde{O}=fc(\tilde{\Phi})$;       // Generate multiple outputs\\
\STATE $\mathcal{L}_{TRR}\gets$Eq.~\ref{eq:totalloss};       // Calculate the loss function\\
\STATE Backpropagation and optimize model parameters $\theta$;
\ENDFOR
\STATE \textbf{return} Trained SNN.
\end{algorithmic}
\end{algorithm}

\subsection{Temporal Reversal Regularization}

The overview of our TRR method is shown in Fig.~\ref{overview}. For temporal/static data, we obtain the temporally reversed data/feature by input/feature temporal reversal, respectively, and finally generate the output $O$ and the temporally reversed output $\hat{O}$ by forward propagation in the SNN. In addition, after the penultimate layer of the SNN, we hybridize the original and temporally reversed spike firing rates using a ``star operation" to obtain the hybrid firing rate $\tilde{\Phi}$, which is passed to the final classification layer to generate the temporally hybridization output $\tilde{O}$. To make the SNN to be insensitive to these perturbations, we use consistency loss and task loss to learn generalized feature representations. The overall objective during training is shown in Eq.~\ref{eq:totalloss}, and the training algorithm is given in Algorithm~\ref{alg:trt_training}. More PyTorch-style pseudocode can be found in the \textbf{Appendix~A}.
\vskip -0.2in
\begin{equation}
\label{eq:totalloss}
\mathcal{L}_{TRR}\!=\!(1-\alpha)\mathcal{L}_{CE}(O,Y) \! + \! \mathcal{L}_{con}(O,\hat{O}) \! + \! \alpha\mathcal{L}_{CE}(\tilde{O},Y).
\end{equation}

During training, our method logically transforms the SNN into a multi-head architecture (exploiting the inherent temporal properties of spiking neurons to produce multiple distinct outputs) to learn generalized representations. During inference, our SNN behaves like vanilla SNNs, generating a single regular prediction without compromising its inference efficiency. In addition, our method is versatile for a variety of tasks, independent of specific architectures and spiking neuron types, providing excellent generalizability.

\vskip -0.2in
\subsection{Theoretical Analysis}
The PAC-Bayesian framework~\cite{10.1007/978-3-540-45167-9_16,NIPS2017_10ce03a1} provides bounds on the expected error of a random predictor, and here we demonstrate the effectiveness of the proposed method in facilitating generalization through PAC-Bayesian theory.

Given a prior distribution $\mathcal{P}$ of network weights $\theta$, $\mathcal{Q}$ is the posterior distribution of $\theta$, and $\mathcal{D}$ is the data distribution, for any $\delta \in (0,1]$, we are able to obtain at least with probability $1-\delta$~\cite{NIPS2017_10ce03a1}:
\vskip -0.2in
\begin{equation}
\mathbb{E}_{\theta \in \mathcal{Q}}[\mathcal{L}_\mathcal{D}(\theta)] \le
\mathbb{E}_{\theta \in \mathcal{Q}}[\hat{\mathcal{L}}_\mathcal{D}(\theta)] 
+
4\sqrt{\frac{KL(\mathcal{Q}||\mathcal{P} + \text{ln} \frac{2m}{\delta})}{m}},
\label{theory1}
\end{equation}
where $\mathcal{L}_\mathcal{D}(\theta)$ is the expected risk, $\hat{\mathcal{L}}_\mathcal{D}(\theta)$ is the empirical risk, and $m$ is the sample size.

\begin{table*}[t]
 \centering
 \scalebox{0.85}{
 \begin{tabular}{ccccccccc}
  \toprule
 Method & TR & FH & CIFAR10 & CIFAR100 & CIFAR10-DVS & DVS-Gesture & ModelNet10 & ModelNet40\\
  \midrule
  Baseline & & & 93.67 & 73.39 & 73.97 & 87.85 & 92.38 & 87.35\\
  +TR & \ding{51} & & $94.18_{+0.51}$ & $74.45_{+1.06}$ & $76.87_{+2.90}$ & $91.32_{+3.47}$ & $93.20_{+0.82}$ & $87.86_{+0.41}$\\
  +FH & & \ding{51} & $94.02_{+0.35}$ & $73.81_{+0.42}$ & $75.33_{+1.36}$ & $90.16_{+2.31}$ & $93.26_{+0.88}$ & $88.16_{+0.81}$\\
  TRR & \ding{51} & \ding{51} &  $\textbf{94.45}_{+0.78}$ & $\textbf{74.85}_{+1.46}$ & $\textbf{77.60}_{+3.63}$ & $\textbf{91.67}_{+3.82}$ & $\textbf{93.45}_{+1.07}$ & $\textbf{88.84}_{+1.49}$\\
  \bottomrule
 \end{tabular}
 }
 \vskip -0.1in
 \caption{Ablation study results (\%) of the proposed method (TR: Temporal Reversal, FH: Feature Hybridization).}
 \label{ablation:vgg}
\vskip -0.22in
\end{table*}

We denote the perturbation (using the input perturbation as an example, the feature perturbation can be obtained in the same way) by $\mathcal{T}$, let the distribution of the data after the perturbation be $\mathcal{D}_\mathcal{T}$. During training, consistency loss pushes the model to align the distribution of outputs to $X$ and $\hat{X}$, i.e., $KL(p||\hat{p})\le \epsilon$, which is equivalent to constraining the model's outputs to perturbations that do not change by more than a minimal value $\epsilon$. Thus, the posterior distribution $\mathcal{Q}_{con}$ optimized by the consistency loss implicitly satisfies robustness to the perturbation $\mathcal{T}$. In addition, the consistency loss can be viewed as an implicit compression of $KL(\mathcal{Q}_{con}||\mathcal{P})$. While the prior $\mathcal{P}$ typically does not take into account perturbation robustness (generalization), the posterior $\mathcal{Q}_{con}$ compresses the parameter space by optimizing the consistency loss $\mathcal{L}_{con}$. So it can be obtained:
\vskip -0.13in
\begin{equation}
KL(\mathcal{Q}_{con}||\mathcal{P}) \propto \mathbb{E}_{\theta \in \mathcal{Q}}[KL(p||\hat{p})].
\label{theory2}
\end{equation}
\vskip -0.03in
By combining Eq.~\ref{theory1} and Eq.~\ref{theory2}, the upper bound of the generalization error based on the consistency loss is
$\mathcal{L}_{con}$. So it can be obtained:
\vskip -0.22in
\begin{equation}
\begin{aligned}
\mathbb{E}_{\theta \in \mathcal{Q}_{con}}[\mathcal{L}_\mathcal{D}(\theta)]
&\le
\mathbb{E}_{\theta \in \mathcal{Q}_{con}}[\hat{\mathcal{L}}_\mathcal{D}(\theta)] \\
&+
4\sqrt{\frac{\mathbb{E}_{\theta \in \mathcal{Q}}KL(p||\hat{p}) + \text{ln} \frac{2m}{\delta})}{m}}.
\label{theory3}
\end{aligned}
\end{equation}
\vskip -0.1in
Eq.~\ref{theory3} shows that optimizing the consistency loss directly reduces the KL divergence term $KL(p||\hat{p})$, which indirectly tightens the upper bound on the generalization error and improves the generalization. In addition, the feature hybrid perturbation is equivalent to increasing the effective number of samples ($m$ in Eq.~\ref{theory3}), which further tightens the upper bound on the generalization error. 

\begin{table*}[!tb]
 \centering
 \tabcolsep=0.02\columnwidth
 \begin{threeparttable}
 \scalebox{0.9}{
 \begin{tabular}{cccccccc}
  \toprule
  Method & Type & Architecture & Spike-driven &  Param (M) & T & Accuracy\\
  \midrule
  RecDis-SNN~\cite{RecDis-SNN}$^{CVPR'22}$ & Surrogate gradient &  ResNet-34 & \ding{51} & 21.79 & 6 & 67.33 \\
  RMP-Loss~\cite{RMPloss}$^{ICCV'23}$ & Surrogate gradient &  ResNet-34 & \ding{51} & 21.79 & 4 & 65.17 \\
  SSCL~\cite{zhang2024enhancing}$^{AAAI'24}$ & Surrogate gradient & ResNet-34 & \ding{51} & 21.79 & 4 & 66.78\\
  TAB~\cite{TAB}$^{ICLR'24}$ & Surrogate gradient & ResNet-34 & \ding{51} & 21.79 & 4 & 67.78\\
  Shortcut~\cite{guo2024take}$^{NeurIPS'24}$ & Surrogate gradient & ResNet-34 & \ding{51} & 21.79 & 4 & 68.14\\
  IMP+LTS~\cite{shen2024rethinking}$^{NeurIPS'24}$ & Surrogate gradient & SEW ResNet-34 & \ding{55} & 25.54 & 4 & 68.90\\
  MPS~\cite{MPS}$^{ICLR'25}$ & Surrogate gradient & SEW ResNet-34 & \ding{55} & 25.54 & 4 & 69.03\\
  MS-ResNet~\cite{MSResNet}$^{TNNLS'24}$ & Surrogate gradient &  MS-ResNet34 & \ding{51} & 21.80 & 6 & 69.42 \\
  GAC-SNN~\cite{qiu2024gated}$^{AAAI'24}$ & Surrogate gradient & MS-ResNet34  & \ding{51} & 21.93 & 4 & 69.77\\
  \hline
  \rowcolor{gray!15}\textbf{TRR (Ours)} & Surrogate gradient & MS-ResNet34  & \ding{51} & 21.93 & 4 & \textbf{70.17}\\
  \hline
  SDT~\cite{Spike_driven_Transformer}$^{NeurIPS'23}$ & Surrogate gradient & SDT Transformer-8-768  & \ding{51} & 66.34 & 2 & 73.06\tnote{$\diamond$}$\thickspace$/74.32\tnote{$\diamond\dag$}\\
  \hline
  \rowcolor{gray!15}\textbf{TRR (Ours)} & Surrogate gradient & SDT Transformer-8-768 & \ding{51} & 66.34 & 2 & \textbf{74.01}/\textbf{74.77}\tnote{$\dag$}\\
  \bottomrule
 \end{tabular}
 }
 \end{threeparttable}
 \vskip -0.1in
 \caption{Comparative results (\%) on ImageNet. $\dag$ denotes an inference resolution of $288\times288$, the default resolution is $224\times224$. $\diamond$ indicates a 2 timestep inference using a publicly available 4 timestep trained checkpoint.}
 \label{com_imagenet}
\vskip -0.1in
\end{table*}

\begin{table*}[!t]
 \centering
 \begin{threeparttable}
 \scalebox{0.9}{
 \begin{tabular}{cccccc}
  \toprule
  Method & Type & Architecture & T & CIFAR10-DVS & DVS-Gesture\\
  \midrule
  NDOT~\cite{NDOT}$^{ICML'24}$ & Forward-in-time & VGG-11 & 10 & 77.50 & -\\
  SLT~\cite{anumasa2024enhancing}$^{AAAI'24}$ & Surrogate gradient &  VGG-9 & 5 & 74.23\tnote{*} & 89.35\tnote{*}\\
  CLIF~\cite{CLIF}$^{ICML'24}$ & Surrogate gradient &  VGG-9 & 5 & 74.97\tnote{*} & 91.55\tnote{*}\\
  SSNN~\cite{SSNN}$^{AAAI'24}$ & Surrogate gradient & VGG-9 & 5 & 73.63 & 90.74\\
  MPS~\cite{MPS}$^{ICLR'25}$ & Surrogate gradient & VGG-9 & 5 & 76.77 & 93.23\\
  \hline
   \multirow{2}{*}{SDT~\cite{Spike_driven_Transformer}$^{NeurIPS'23}$} & \multirow{2}{*}{Surrogate gradient} & \multirow{2}{*}{SDT Transformer-2-256} & 5 &  72.53\tnote{*} & 94.33\tnote{*}\\& & & 16 & 80.00 & \textbf{99.30}\\
   \hline
  QKFormer~\cite{QKFormer}$^{NeurIPS'24}$ & Surrogate gradient & HST-2-256 & 16 & 84.00 & 98.60 \\
  \hline
  \rowcolor{gray!15} &  & SDT Transformer-2-256 & 5 & 75.55 & 96.88\\
  \rowcolor{gray!15} & & HST-2-256 & 16 & \textbf{84.30} & 99.03\\ 
  \rowcolor{gray!15} \multirow{-3}{*}{\textbf{TRR (Ours)}}  & \multirow{-3}{*}{Surrogate gradient} & VGG-9 & 5 & 77.60 & 91.67\\
  \hline
   \rowcolor{orange!10}\textbf{TRR (Ours)} + CLIF~\cite{CLIF} & Surrogate gradient &  VGG-9 & 5 & 78.20 & 94.10\\
  \bottomrule
 \end{tabular}
 }
 \end{threeparttable}
 \vskip -0.1in
 \caption{Comparative results (\%) on neuromorphic datasets. * denotes self-implementation results with open-source code.}
 \label{com_dvs}
 \vskip -0.25in
\end{table*}
\vskip -0.1in
\section{Experiments}

To confirm the effectiveness and generalizability of our method, we conduct experiments on the tasks of static object recognition, neuromorphic object/action recognition, and 3D point cloud classification using VGG-9~\cite{SSNN}, MS-ResNet~\cite{MSResNet}, Spike-driven Transformer~\cite{Spike_driven_Transformer}, QKFormer~\cite{QKFormer}, PointNet~\cite{Qi_2017_CVPR}, and PointNet++~\cite{NIPS2017_d8bf84be} architectures. If not specified, the timestep was 5 for neuromorphic datasets and 2 for static datasets, reflecting the performance of the SNN at low latencies. The experimental details can be found in \textbf{Appendix B}.

\vskip -0.05in
\subsection{Ablation Studies}
\label{ablationstudy}
\vskip -0.05in

\begin{figure}[!t]
\centering
\includegraphics[width=0.48\textwidth]{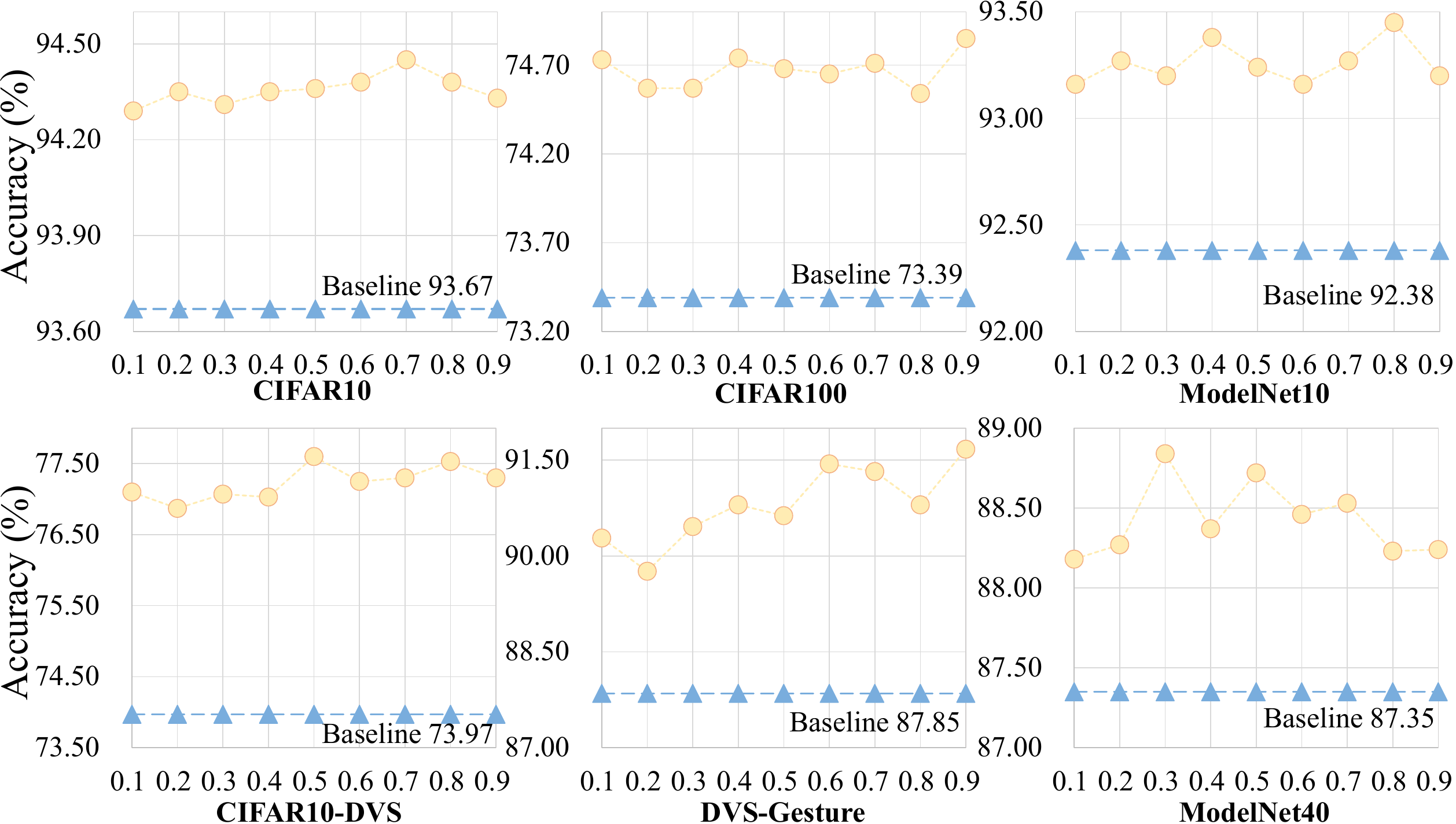}
\vskip -0.1in
\caption{Influence of the balance coefficient $\alpha$ on the performance. A larger $\alpha$ indicates stronger perturbation regularization. As a whole, our method is insensitive to $\alpha$ and consistently outperforms the baseline.}
\label{ablation:alpha}
\vskip -0.25in
\end{figure}

\subsubsection{Hyperparameter Sensitivity Analysis}
In Fig.~\ref{ablation:alpha}, we have experimentally studied the influence of the balance coefficient $\alpha$ on the performance. The influence of $\alpha$ is most significant for the DVS-Gesture, where larger values of $\alpha$ yield obviously better results, due to the stronger regularization of the perturbations at this point, which effectively mitigates the overfitting of the model. Overall, $\alpha$ leads to only slight fluctuations in the performance of the SNN, while consistently outperforming the baseline, indicating that our method is not sensitive to $\alpha$. We set the value of $\alpha$ in later experiments based on the performance peaks in Fig.~\ref{ablation:alpha}.

\vskip -0.1in
\subsubsection{Comparison to Baseline} The ablation studies for our method are shown in Tab.~\ref{ablation:vgg}, where the PointNet was used for ModelNet10/40 and VGG-9 for the other datasets. Experimental results show that using temporal reversal and feature hybridization alone improves the performance of the baseline SNN, and the maximum performance gain is achieved when training with both together. In particular, TRR was most effective on the neuromorphic datasets CIFAR10-DVS and DVS-Gesture, confirming that our method is able to mitigate overfitting and is not affected by temporal order reversal. In particular, high-precision, low-latency recognition of neuromorphic objects by SNNs contributes to the real-world deployment of neuromorphic software-hardware integration by maximizing its performance and speed advantages~\cite{yao2024spike}. Additional ablation studies using the MS-ResNet and Spike-driven Transformer architectures in \textbf{Appendix~C.1} show the same conclusions.

\begin{figure}[!t]
\centering
\includegraphics[width=0.5\textwidth]{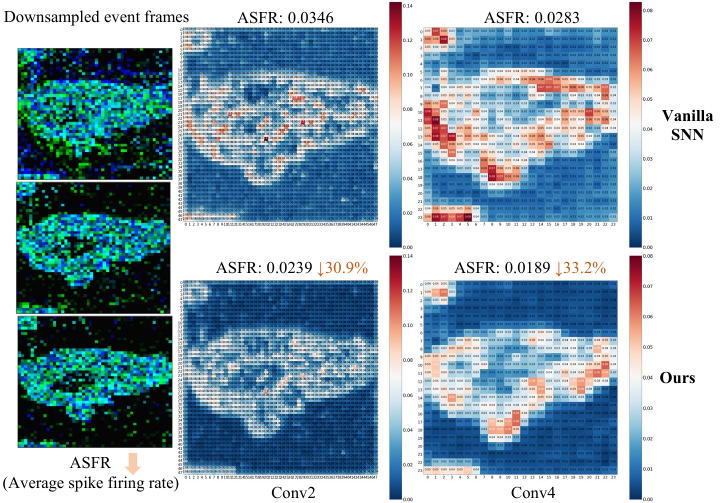}
\vskip -0.1in
\caption{Visualization of ASFR. Our method has a lower ASFR than the baseline, favoring low-power deployment.}
\label{visASFR}
\vskip -0.15in
\end{figure}

\vskip -0.2in
\subsubsection{Average Spike Firing Rate Visualization}

We have visualized the average spike firing rate (ASFR) of the first two stages in VGG-9 on CIFAR10-DVS in Fig.~\ref{visASFR}. Compared to the vanilla SNN, our method not only achieves better performance but also has a lower ASFR (ASFR is positively correlated with the power overhead during deployment~\cite{Spike_driven_Transformer}), indicating that our method is more suitable for training low-energy SNNs to be deployed on edge devices. In \textbf{Appendix D}, we show additional visualizations (across CIFAR10-DVS and DVS-Gesture) that illustrate that the lower ASFR gains from TRR are consistently observed.

\begin{table*}[tb]
 \centering
 \tabcolsep=0.04\columnwidth
 \begin{threeparttable}
 \scalebox{0.9}{
 \begin{tabular}{cccccc}
  \toprule
  Method & Type & Architecture & T & ModelNet10 & ModelNet40\\
  \midrule
  PointNet~\cite{Qi_2017_CVPR}$^{CVPR'17}$ & ANN & PointNet & - & 93.31\tnote{*} & 89.46\tnote{*} \\
  PointNet++~\cite{NIPS2017_d8bf84be}$^{NeurIPS'17}$ & ANN & PointNet++ & - & 95.50\tnote{*} & 92.16\tnote{*} \\
  Converted SNN~\cite{Lan_2023_ICCV}$^{ICCV'23}$ & SNN & PointNet & 16 & 92.75 & 88.17\\
  Spiking PointNet~\cite{Spiking_PointNet}$^{NeurIPS'23}$ & SNN & PointNet & 2 & 92.98\tnote{*} & 87.58\tnote{*} \\
  P2SResLNet~\cite{wu2024point}$^{AAAI'24}$ & SNN & P2SResLNet & 1 & - & 89.20\\
  \hline
  \rowcolor{gray!15}& &PointNet &  2 & 93.45 & 88.84\\
  \cline{3-6}
  \rowcolor{gray!15}&&  &  2 & \textbf{93.97} & \textbf{90.57} \\
  \rowcolor{gray!15}\multirow{-3}{*}{\textbf{TRR (Ours)}} & \multirow{-3}{*}{SNN}& \multirow{-2}{*}{PointNet++} &  1 & 93.31\tnote{$\diamond$} & 89.65\tnote{$\diamond$}\\
  \bottomrule
 \end{tabular}
 }
 \end{threeparttable}
 \vskip -0.1in
 \caption{Comparative results (\%) on point cloud classification. * self-implementation. $\diamond$ training: $T=2$, inference: $T=1$.}
 \label{com_point}
 \vskip -0.18in
\end{table*}

\vskip -0.2in
\subsection{Comparison with Existing Methods}

\textbf{Static Object Recognition.} The comparative results of TRR with other methods on ImageNet are shown in Tab.~\ref{com_imagenet}. Using the MS-ResNet34 architecture, TRR achieves an accuracy of 70.17\%, significantly outperforming other methods using the same architecture. We also applied our method to the SDT~\cite{Spike_driven_Transformer} architecture and achieved an accuracy of 74.77\% at two timesteps, surpassing their official published model. This indicates that our method can consistently improve the performance of purpose-built architectures. The comparative results for the CIFAR10/100 dataset are shown in \textbf{Appendix~C.3}.

\textbf{Neuromorphic Recognition.} In Tab.~\ref{com_dvs}, TRR achieved an accuracy of 84.30\% on CIFA10-DVS, surpassing the other methods. When compared under the same conditions on DVS-Gesture, our method consistently outperforms SDT~\cite{Spike_driven_Transformer} and QKFormer~\cite{QKFormer}, confirming the generalizability of our method. When vanilla LIF neurons were replaced with CLIF~\cite{CLIF}, the performance of the VGG-9 TRR was further improved, suggesting that the TRR is able to seamlessly integrate with other neurons.

\textbf{3D Point Cloud Classification.} Compared to 2D object recognition, 3D point cloud classification is more challenging and there are few available SNN methods. As shown in Tab.~\ref{com_point}, our method again achieves the optimal SNN performance on the point cloud classification task. P2SResLNet~\cite{wu2024point} achieves 89.20\% accuracy on ModelNet40 with computationally expensive 3D spiking residual blocks, while we outperform it by 0.45\% at the same timestep using the lightweight PointNet++ architecture.

\subsection{Influence of Feature Reversal Location}

For static data, TRR temporally reverses the encoded spikes. We explored the influence of temporal reversal at different locations using VGG-9 on CIFAR10/100 (defaulted to stage~1), and the results are shown in Tab.~\ref{ablation:location}. The later the location of the feature temporal reversal, the smaller the performance gain of the TRR, but it still outperforms the baseline model. We attribute this to the fact that the more advanced the feature reversal location, the more the reversed output differs from the original output, i.e., the stronger the perturbation, and thus the better it promotes generalization.

\begin{table}[t]
 \centering
 \scalebox{0.9}{
 \begin{tabular}{cccc|c}
  \toprule
Reversal location & Stage 1 & Stage 2 & Stage 3 & Baseline\\
  \midrule
  CIFAR10 & 94.45 & 94.34 & 94.06 & 93.67\\
  CIFAR100 & 74.85 & 74.67 & 74.40 & 73.39\\
  \bottomrule
 \end{tabular}
 }
 \vskip -0.1in
 \caption{Influence of feature temporal reversal location (\%). Stage~$i$ denotes feature temporal reversal after the $i$th stage. The further ahead of the location, the greater the performance gain, and it consistently outperforms the baseline.}
 \label{ablation:location}
 \vskip -0.1in
\end{table}

\begin{table}[t]
 \centering
 \scalebox{0.9}{
 \begin{tabular}{cccc}
  \toprule
  Method & T & CIFAR10-DVS & DVS-Gesture\\
  \midrule
  EventAugment~\cite{EventAugment} & 20 &- & 96.25\\
  EventDrop~\cite{EventDrop} & 20 & - & 93.75\\
  NDA~\cite{NDA} & 5 & 76.47 & 90.16\\
  \hline
  TR & 5 & 76.87 & 91.32\\
  \rowcolor{gray!15}TRR & 5 & 77.60 & 91.67\\
  \hline
  TRR(\ding{51}NDA) & 5 & 78.20 & 92.48\\
  \rowcolor{gray!15}TRR(\ding{59}NDA) & 5 & 78.77 & 94.33\\
  \bottomrule
 \end{tabular}
 }
 \vskip -0.1in
 \caption{Comparison and extension with neuromorphic data augmentation (\%). \ding{51}: NDA is used before TRR. \ding{59}: NDA replaces temporal reversal but keeps consistency and perturbation loss.}
 \label{com_dvsaug}
 \vskip -0.2in
\end{table}

\subsection{Extension with Data Augmentation}

In Tab.~\ref{com_dvsaug}, we compare TRR (VGG-9) with other neuromorphic data augmentation methods. We achieved better performance than NDA~\cite{NDA} using only temporal reversal. EventAugment~\cite{EventAugment} and EventDrop~\cite{EventDrop} achieved 96.25\% and 93.75\% accuracy on DVS-Gesture with $T=20$, respectively, and TRR achieved 91.67\% accuracy with only five timesteps. In particular, the TRR with SDT Transformer architecture achieves an accuracy of 96.88\% for $T=5$ (Tab.~\ref{com_dvs}), exceeding both.

In addition, TRR can be further extended by incorporating data augmentation. We explore two extensions: (1) NDA followed by TRR (denoted as \ding{51}), and (2) replacing the input temporal reversal with NDA (denoted as \ding{59}), preserving the consistency loss (Eq.~\ref{TRloss}) and the perturbation loss (Eq.~\ref{FHloss}). The results in Tab.~\ref{com_dvsaug} show that both extensions lead to further enhancement of the performance. In particular, learning the consistency between the augmented data and the original input yields larger performance gains, similar to self-supervised learning of the generalized features of two augmented views~\cite{pmlr-v119-lee20c,pmlr-v119-chen20j}, demonstrating the superior extensibility of TRR.

\subsection{Comparison of Reversal and Shuffle}

In addition to temporal reversal, we also explore the influence of random temporal shuffling in Tab.~\ref{com_shuffle}. The results show that random shuffling also mitigates overfitting, but completely destroys the temporal dimension, making it difficult for SNNs to extract coherent temporal features, and thus performs inferiorly to temporal reversal.

\begin{table}[!t]
 \centering
 \tabcolsep=0.025\columnwidth
 \scalebox{0.9}{
 \begin{tabular}{ccc}
  \toprule
  Method  & CIFAR10-DVS (\%) & DVS-Gesture (\%) \\
  \midrule
  Vanilla SNN & 73.97 & 87.85\\
  \hline
  TRR-shuffle &  76.73 & 91.09\\
  \rowcolor{gray!15}TRR-reversal &  77.60 & 91.67\\
  \bottomrule
 \end{tabular}
 }
 \vskip -0.1in
 \caption{Comparison of random shuffle timestep and reversal.}
 \label{com_shuffle}
 \vskip -0.15in
\end{table}

\vspace{-0.05in}
\subsection{Enhanced Adversarial Robustness from TRR}
\label{robust}
\vspace{-0.05in}
Theoretical analysis shows that TRR promotes robust generalization, and here we investigate whether this robust generalization improves adversarial robustness. We use VGG-11 to conduct experiments on CIFAR100, and the adversarial methods include Gaussian noise (GN) and commonly used adversarial attacks (FGSM, PGD with random start, BIM, and CW), with the same experimental details as~\cite{xu2024feelsnn}.

\begin{table}[tb]
 \centering
 \tabcolsep=0.01\columnwidth
 \begin{threeparttable}
 \scalebox{0.94}{
 \begin{tabular}{c|cccccc}
  \toprule
  Method  & Clean & GN & FGSM & PGD & BIM & CW\\
  \midrule
  FEEL-SNN~\cite{xu2024feelsnn} & 63.95 & 61.97 & 9.87 & 2.13 & 1.93 & 6.21\\
  AT~\cite{goodfellow2014explaining} &  67.97 & 67.47 & 17.55 & 9.52 & 8.91 & 20.23\\
  RAT~\cite{ding2022snnrat} & 69.99 & 69.06 & 19.00 & 9.11 & 8.46 & 22.59\\
  \hline
  \rowcolor{gray!15}\textbf{TRR} & \textbf{70.11} & \textbf{69.44} & 18.60 & \textbf{9.80} & \textbf{9.07} & \textbf{25.33}\\
  \bottomrule
 \end{tabular}
 }
 \end{threeparttable}
 \vskip -0.1in
 \caption{TRR can improve the adversarial robustness of the model.}
 \label{com_robust}
 \vskip -0.25in
\end{table}

We combine TRR with RAT~\cite{ding2022snnrat}, and the results in Tab.~\ref{com_robust} show that the adversarial robustness of the model is further improved, especially for the CW attack, the robust accuracy of the model is improved by 2.74\%. This suggests that the perturbations introduced during training make the SNN more robust to malicious attacks, which facilitates its security when deployed on low-power edge devices (which is the ultimate strength of SNNs).

\vspace{-0.1in}
\section{Conclusion}
\vspace{-0.05in}

In this paper, we propose the temporal reversal regularization, which significantly improves the generalization performance of SNNs through temporal reversal and feature hybridization for perturbation during training. We show theoretically that our method can tighten the upper bound on the generalization error, and verify its effectiveness, generality, and adversarial robustness through extensive experiments. We expect our work to unleash the potential of high-performance SNNs.

{
    \small
    \bibliographystyle{ieeenat_fullname}
    \bibliography{main}
}

\newpage
\section*{Appendix Description}

\begin{itemize}
    \item The PyTorch-style pseudocode of the method proposed in this paper is provided in Appendix~\ref{app_pseudocode} for reproduction.
    \item The experimental details of this paper are presented in Appendix~\ref{appendix_b}.
    \item Due to space limitations in the main paper, additional experiments are presented in Appendix~\ref{appendix_c}. Ablation experiments using MS-ResNet and Spike-driven Transformer architectures are presented in~\ref{appendix_c1} the influence of spike hybridization and spike firing rate hybridization are compared in~\ref{appendix_c2}; and the results of the proposed method compared to other methods on the CIFAR10/100 dataset are shown in~\ref{appendix_cifar}.
    \item Visualizations are shown in Appendix~\ref{app_vis}, including the spike firing rate perturbations in~\ref{appendix_d1} and the average spike firing rate in~\ref{appendix_d2}.
\end{itemize}

\newpage
\appendix
\section{Appendix A PyTorch-style Pseudocode Implementation}
\label{app_pseudocode}

The PyTorch-style pseudocode for temporal reversal and spike firing rate hybridization is presented in Algorithm~\ref{alg: trcode} and Algorithm~\ref{alg: sfrcode} to facilitate the understanding and reproduction of our TRR method.

\begin{algorithm}[h]\small
    \caption{PyTorch-style code for temporal reversal}
    \label{alg: trcode}
    \definecolor{codeblue}{rgb}{0.25,0.5,0.5}
    \definecolor{codepink}{rgb}{1,0.5,0.5}
    \definecolor{codedark}{rgb}{1,0.7,0.8}
    \lstset{
        backgroundcolor=\color{white},
        basicstyle=\fontsize{7.2pt}{7.2pt}\ttfamily\selectfont,
        columns=fullflexible,
        breaklines=true,
        captionpos=b,
        commentstyle=\fontsize{8pt}{8pt}\color{codeblue},
        keywordstyle=\fontsize{8.0pt}{8.0pt}\color{codepink},
        emph={DemoNet, Blk}, %
        emphstyle=\color{purple}, %
    }
    {\small
    \begin{lstlisting}[language=python]
# x: input data or encoded spikes.
# x.shape: [T,B,C,H,W] 
# reversed_x: temporally reversed data or spikes.
# reversed_x.shape: [T,B,C,H,W]
def temporal_reversal(x):
    T,B,C,H,W = x.shape
    reversed_x = torch.zeros(x.shape)
    for t in range(T):
        reversed_x[t] = x[T+1-t]
    return reversed_x
    \end{lstlisting}
    }
\end{algorithm}

\begin{algorithm}[h]\small
    \caption{PyTorch-style code for spike firing rate hybridization}
    \label{alg: sfrcode}
    \definecolor{codeblue}{rgb}{0.25,0.5,0.5}
    \definecolor{codepink}{rgb}{1,0.5,0.5}
    \definecolor{codedark}{rgb}{1,0.7,0.8}
    \lstset{
        backgroundcolor=\color{white},
        basicstyle=\fontsize{7.2pt}{7.2pt}\ttfamily\selectfont,
        columns=fullflexible,
        breaklines=true,
        captionpos=b,
        commentstyle=\fontsize{8pt}{8pt}\color{codeblue},
        keywordstyle=\fontsize{8.0pt}{8.0pt}\color{codepink},
        emph={DemoNet, Blk}, %
        emphstyle=\color{purple}, %
    }
    {\small
    \begin{lstlisting}[language=python]

# feat: spike feature.
# reversed_feat: temporally reversed spike feature.
# feat.shape=reversed_feat.shape: [T,B,C,H,W].
# hybridized_fr: hybridized spike firing rate.
# hybridized_fr.shape: [B,C,H,W]
def firing_rate_hybridization(feat, reversed_feat):
    T,B,C,H,W = feat.shape
    fr = torch.zeros((B,C,H,W))
    reversed_fr = torch.zeros(fr.shape)
    for t in range(T):
        fr += feat[t]
        reversed_fr += reversed_feat[t]
    fr /= T
    reversed_fr /= T
    hybridized_fr = fr * reversed_fr
    return hybridized_fr
    \end{lstlisting}
    }
\end{algorithm}

\section{Appendix B Experimental Details}
\label{appendix_b}

\begin{table*}[!h]
 \centering
 \begin{tabular}{cccc}
  \toprule
  Stage & VGG-9 & MS-ResNet18 & MS-ResNet34\\
  \midrule
  0 & - & Conv($3 \times 3@64$) & Conv($7 \times 7@64$)\\
  \hline
  1  &  \makecell{Conv($3 \times 3$@64) \\ Conv($3 \times 3$@128)} &
  \makecell{
  $\left(
 	    \begin{array}{cc}  
 			 \makecell{\text{Conv}(3 \times 3@128) \\ \text{Conv}(3 \times 3@128)}
        \end{array}
    \right)\times 3
  $}
  &
  \makecell{
  $\left(
 	    \begin{array}{cc}  
 			 \makecell{\text{Conv}(3 \times 3@64) \\ \text{Conv}(3 \times 3@64)}
        \end{array}
    \right)\times 2
  $}
  \\
  \hline
    & average pool(stride=2) & - & -\\
  \hline
  2  & \makecell{Conv($3 \times 3$@256) \\ Conv($3 \times 3$@256)} & 
  $\left(
 	    \begin{array}{cc}  
 			 \makecell{\text{Conv}(3 \times 3@256) \\ \text{Conv}(3 \times 3@256)}
        \end{array}
    \right)\times 3
  $
  &
  \makecell{
  $\left(
 	    \begin{array}{cc}  
 			 \makecell{\text{Conv}(3 \times 3@128) \\ \text{Conv}(3 \times 3@128)}
        \end{array}
    \right)\times 4
  $}
  \\
  \hline
    & average pool(stride=2) & - & -\\
  \hline
  3  & \makecell{Conv($3 \times 3$@512) \\ Conv($3 \times 3$@512)} & $\left(
 	\begin{array}{cc}  
 			 \makecell{\text{Conv}(3 \times 3@512) \\ \text{Conv}(3 \times 3@512)}
 \end{array}
 \right)\times 2$
 &
 \makecell{
  $\left(
 	    \begin{array}{cc}  
 			 \makecell{\text{Conv}(3 \times 3@256) \\ \text{Conv}(3 \times 3@256)}
        \end{array}
    \right)\times 6
  $}
 \\
 
  \hline
    & average pool(stride=2) & - & -\\
  \hline
  4
      & \makecell{Conv($3 \times 3$@512) \\ Conv($3 \times 3$@512)} &
-
&
$\left(
 	\begin{array}{cc}  
 			 \makecell{\text{Conv}(3 \times 3@512) \\ \text{Conv}(3 \times 3@512)}
 \end{array}
 \right)\times 3$
\\
  \hline
    \multicolumn{4}{c}{global average pool, fc}\\
  \bottomrule
 \end{tabular}
  \caption{Structures of VGG-9, MS-ResNet18, and MS-ResNet34, where fc denotes the fully connected layer.}
 \label{model}
\end{table*}

\subsection{Tasks and Datasets}
We validate the effectiveness and versatility of the proposed method on a variety of tasks and datasets described below.
\subsubsection{Static Object Recognition}
For the static object recognition task, we use the CIFAR10/100~\cite{CIFAR} and ImageNet~\cite{ImageNet} datasets. 

CIFAR10 contains 50,000 training images and 10,000 test images, each $32\times32$ in size, covering ten types of objects. The CIFAR100 dataset has the same number of training samples, test samples, and image sizes as CIFAR10, but includes one hundred objects with higher recognition difficulty. 

The ImageNet dataset of 1.2 million training images, 50,000 validation images, and 150,000 test images with 1,000 categories is the most challenging object recognition benchmark. For the ImageNet dataset, we unify the images to a $224\times224$ size during training and testing, and evaluate the performance of our method on the test set.

For CIFAR10 and CIFAR100 data, we preprocessed them using standard data augmentation strategies: random cropping, horizontal flipping, and normalization. We also use the autoaugment strategy~\cite{cubuk2019autoaugment} for CIFAR10. For ImageNet data, we use the same data augmentation strategies such as random augmentation and mixup as in~\cite{Spike_driven_Transformer}. Please refer to~\cite{Spike_driven_Transformer} for specific augmentation details.

\subsubsection{Neuromorphic Object Recognition}
For neuromorphic object recognition, we use the CIFAR10-DVS dataset~\cite{CIFAR10-DVS}, which is the neuromorphic version of the CIFAR10 dataset. The CIFAR10-DVS dataset has 10,000 samples for a total of 10 object classes, and the dimension of each sample is $[t,p,x,y]$, where $t$ is the timestamp, $p$ is the polarity of the intensity change of the corresponding pixel, and $x$ and $y$ are the spatial coordinates of the pixel point, respectively. The spatial size of each sample in CIFAR10-DVS is $128\times128$, which we downsampled to $48\times48$ resolution before inputting to the SNN. Additionally, due to the high temporal resolution of the neuromorphic dataset, we integrate a neuromorphic sample into $T$ event frames [$T,p,x,y]$ using the SpikingJelly framework~\cite{SpikingJelly} to match the timestep $T$ of the SNN. For each training, we randomly divide 90\% of the data as the training set and test on the remaining 10\% of the data, which is by far the most common strategy~\cite{SSNN}.

\subsubsection{Neuromorphic Action Recognition}
The DVS-Gesture~\cite{DVS-Gesture} dataset contains neuromorphic data for 11 hand gestures with 1176 training samples and 288 test samples. The dimension of each sample is [$T,p,x,y]$, and we downsample its spatial resolution from $128\times128$ to $48\times48$ before feeding the samples into the SNN. The pre-processing of the DVS-Gesture data is the same as in CIFAR10-DVS, which also utilizes the SpikingJelly framework to obtain the event frame [$T,p,x,y]$ by integrating it by timestep.

\subsubsection{3D Point Cloud Classification}

For the 3D point cloud classification task, we use the ModelNet10 and ModelNet40 datasets~\cite{Wu_2015_CVPR}. The ModelNet10 dataset contains 4,899 3D models in ten different categories, such as tables, chairs, bathtubs, and guitars. The ModelNet40 dataset contains 12,311 3D models in 40 different categories, making it even more challenging.

For the preprocessing of ModelNet10/40 data, we followed~\cite{Spiking_PointNet}: uniformly sampling 1024 points on mesh faces based on the area of the grid surface and normalizing it to the unit sphere. These data of length 1024 are repeatedly fed into the SNN at each timestep.

\subsection{Implementation Details}

In this paper, all experiments are based on the PyTorch package running on both Nvidia RTX 4090 and 3090 GPUs. For both static object recognition and neuromorphic datasets, we use three architectures, VGG-9~\cite{SSNN}, MS-ResNet18~\cite{MSResNet}, and Spike-driven Transformer~\cite{Spike_driven_Transformer}. For the VGG-9 and MS-ResNet architectures, we follow the training strategy of~\cite{SSNN}: train the model with an initial learning rate of 0.1 for 100 epochs, reducing it by a factor of ten every 30 epochs. A stochastic gradient descent optimizer with a momentum of 0.9 and a batch size of 64 was used. The weight decays for the static and neuromorphic datasets are 1e-4 and 1e-3, respectively. We used the LIF neuron model with a firing threshold $\vartheta$ of 1.0 and a membrane potential time constant $\tau$ of 2.0.

When using the Spike-driven Transformer architecture, we follow the training strategy of the original paper~\cite{Spike_driven_Transformer}: 300 epochs on static datasets and 200 epochs on neuromorphic datasets; the network structures used in CIFAR-10, CIFAR-100, ImageNet, CIFAR10-DVS, and DVS-Gesture are: spike-driven Transformer-2-512, spike-driven Transformer-2-512, Spiking Transformer-8-768, spike-driven Transformer-2-256, spike-driven Transformer-2-256. See~\cite{Spike_driven_Transformer} for more details on training.

When using the QKFormer architecture, we use the training strategy and publicly available code of its original paper, see~\cite{QKFormer} for a detailed setup.

For the point cloud classification task, we use the Spiking PointNet~\cite{Spiking_PointNet} and PointNet++~\cite{NIPS2017_d8bf84be} architectures and the training strategy follows~\cite{Spiking_PointNet}: The initial learning rate was set to 0.001 and degraded by 0.7 every 20 epochs for a total of 200 epochs of training using the Adam optimizer. See~\cite{Spiking_PointNet} for more details on training.

To reduce the influence of randomness, we repeated all our experiments three times, and the average results are reported in the paper. However, for the ImageNet dataset, which has a huge training overhead, and for adversarial robustness experiments, we still report single trial results.

\subsection{Network Architectures}
The VGG-9 network consists of eight convolutional-spiking layers and a fully connected layer for classification. MS-ResNet contains multiple contiguous residual blocks and uses identity connections between the membrane potentials of the spiking neurons. We made minor modifications to the MS-ResNet18 architecture in the original paper~\cite{MSResNet} according to~\cite{qiu2024gated} (the $7\times7$ convolution kernel of the first convolution was replaced by $3\times3$ and stride was set to 1), and kept the original MS-ResNet34 architecture~\cite{MSResNet}. In addition, when using MS-ResNet34 for inference on ImageNet, we use the Gated Attention Coding method~\cite{qiu2024gated}. The specific architectural details are shown in Table~\ref{model}. 

\begin{table*}[t]
 \centering
 \begin{tabular}{ccccc}
  \toprule
 Method & CIFAR10 & CIFAR100 & CIFAR10-DVS & DVS-Gesture\\
  \midrule
  Baseline & 94.69 & 75.45 & 66.40 & 89.35\\
  +TR & $95.01_{+0.32}$ & $75.97_{+0.52}$ & $73.83_{+7.43}$ & $91.78_{+2.43}$\\
  +FH & $94.89_{+0.20}$ & $75.95_{+0.50}$ & $70.73_{+4.33}$ & $91.55_{+2.20}$\\
  TRR & $\textbf{95.13}_{+0.44}$ & $\textbf{76.14}_{+0.69}$ & $\textbf{74.60}_{+8.20}$ & $\textbf{92.82}_{+3.47}$ \\
  \bottomrule
 \end{tabular}
 \vskip -0.1in
 \caption{Ablation results (\%) of the proposed method using the MS-ResNet18 architecture.}
 \label{ablation:resnet}
\end{table*}

\begin{table*}[t]
 \centering
 \begin{tabular}{ccccc}
  \toprule
 Method & CIFAR10 & CIFAR100 & CIFAR10-DVS & DVS-Gesture\\
  \midrule
  Baseline & 94.91 & 77.63 & 72.53 & 94.33\\
  +TR & $95.45_{+0.54}$  & $78.86_{+1.23}$ & $75.27_{+2.74}$ & $96.18_{+1.85}$\\
  +FH & $94.62_{-0.29}$ & $76.33_{-1.30}$ & $73.90_{+1.37}$ & $95.49_{+1.16}$\\
  TRR & $\textbf{95.61}_{+0.70}$ &  $\textbf{79.43}_{+1.80}$ & $\textbf{75.55}_{+3.02}$ & $\textbf{96.88}_{+2.55}$ \\
  \bottomrule
 \end{tabular}
 \vskip -0.1in
 \caption{Ablation results (\%) of the proposed method using the Spike-driven Transformer architecture.}
 \label{ablation:sdt}
\end{table*}

\subsection{Feature Reversal Location}

For static data, we temporally reverse the spike features taking advantage of the inherent temporal properties of the SNN. For the VGG-9 network, we consider the first two convolutional-spiking layers as the spike encoding module from which the spike features are temporally reversed. For the MS-ResNet network, we consider the first convolutional-spiking layer as the spike encoding module that produces temporally reversed features before the residual block. For the Spike-driven Transformer network, we temporally reverse the spike features generated after patch embedding module. For the Spiking PointNet network, we consider the input transformation within it as the spike encoding module, where the spike features are temporally reversed.

\subsection{Details of Reproduction of Existing Methods}
For a fair comparison with existing methods, the methods in~\cite{Spike_driven_Transformer},~\cite{MSResNet},~\cite{Spiking_PointNet},~\cite{TAB}, and~\cite{anumasa2024enhancing} are reproduced in this paper.

Spike-driven Transformer~\cite{Spike_driven_Transformer}: We implement Spike-driven Transformer using the official code provided in the original paper, keeping the network structure and hyperparameters such as the learning rate unchanged.

MS-ResNet~\cite{MSResNet}: The MS-ResNet18 and MS-ResNet34 architectures we used are shown in Table~\ref{model}; we trained MS-ResNet18 with the same training strategy as VGG-9, and when using MS-ResNet34 we used the training strategy in~\cite{Spike_driven_Transformer}.

Spiking PointNet~\cite{Spiking_PointNet}: We implement Spiking PointNet using the official code provided in the original paper, keeping the network structure and hyperparameters such as the learning rate unchanged.

Temporal Accumulated Batch Normalization~\cite{TAB}: We use the Temporal Accumulated Batch Normalization (TAB) layer to replace the vanilla BN layer in VGG-9, and the other training strategies are consistent with our experiments. We implement the TAB layer according to the officially released code~\cite{TAB}.

Stochastic Latency Training~\cite{anumasa2024enhancing}: When reproducing Stochastic Latency Training (SLT), we use the VGG-9 network architecture and keep the training parameters consistent with our experimental settings. During training, we follow the SLT algorithm to randomly sample the timestep for training, and the timestep for inference is 5. We set the range of timesteps during training to $[1,5]$ and $[1,10]$, and achieved average accuracies of 74.23\%, 89.35\% ($[1,5]$) and 75.00\%, 91.44\% ($[1,10]$) on CIFAR10-DVS and DVS-Gesture, respectively. To ensure a fair comparison, we present results for training timesteps ranging from $[1,5]$ in Table 4. It is worth noting that our method still achieves better performance even when compared to SLT with a training timestep range of $[1,10]$.

\begin{table*}[!t]
 \centering
  \tabcolsep=0.04\columnwidth
 \begin{threeparttable}
 \scalebox{0.9}{
 \begin{tabular}{cccccc}
  \toprule
  Method & Type & Architecture & T & CIFAR10 & CIFAR100\\
  \midrule
  RMP-Loss~\cite{RMPloss}$^{ICCV'23}$ & Surrogate gradient &  VGG-16 & 10 & 94.39 & 73.30\\
  CLIF~\cite{CLIF}$^{ICML'24}$ & Surrogate gradient &  ResNet-18 & 4 & 94.89 & 77.00\\
  SSCL~\cite{zhang2024enhancing}$^{AAAI'24}$ & Surrogate gradient & ResNet-20 & 2 & 93.40 & 69.81\\
  NDOT~\cite{NDOT}$^{ICML'24}$ & Forward-in-time & VGG-11 & 2 & 94.44 & 75.27\\
  MS-ResNet~\cite{MSResNet}$^{TNNLS'24}$ & Surrogate gradient &  MS-ResNet18 & 2 & 94.69\tnote{*} & 73.84\tnote{*}\\
  TAB~\cite{TAB}$^{ICLR'24}$ & Surrogate gradient &  ResNet-19 & 2 & 94.73 & 76.31\\
  SLT-TET~\cite{anumasa2024enhancing}$^{AAAI'24}$ & Surrogate gradient &  ResNet-19 & 2 & 94.96 & 73.77\\
  \hline
  \rowcolor{gray!15}& & VGG-9 & 2 & 94.45 & 74.85\\ 
   \rowcolor{gray!15}\multirow{-2}{*}{\textbf{TRR (Ours)}} & \multirow{-2}{*}{Surrogate gradient}& MS-ResNet18 & 2 & 95.13 & 76.14\\
  \hline
  Spikformer~\cite{Spikformer}$^{ICLR'23}$ & Surrogate gradient & Spiking Transformer-4-256 & 4 & 93.94 & 75.96\\
  \hline
  \multirow{2}{*}{SDT~\cite{Spike_driven_Transformer}$^{NeurIPS'23}$} & \multirow{2}{*}{Surrogate gradient} & \multirow{2}{*}{Spiking Transformer-2-512} & 2 & 94.91\tnote{*} & 77.63\tnote{*} \\
  & & & 4 & 95.60 & 78.40\\
  \hline
  \rowcolor{gray!15} \textbf{TRR (Ours)} & Surrogate gradient & Spiking Transformer-2-512 & 2 & 95.61 & 79.43\\
  \hline
  SNN-ViT~\cite{SNN-ViT}$^{ICLR'25}$ & Surrogate gradient & SNN-ViT & 4 & 96.10 & 80.10\\
  QKFormer~\cite{QKFormer}$^{NeurIPS'24}$ & Surrogate gradient & HST-4-384 & 4 & 96.18 & 81.15\\
  \hline
   \rowcolor{gray!15}\textbf{TRR (Ours)} & Surrogate gradient &  HST-4-384 & 4 & \textbf{96.71} & \textbf{81.65}\\
  \bottomrule
 \end{tabular}
 }
 \end{threeparttable}
 \vskip -0.1in
 \caption{Comparative results (\%) on static datasets. * denotes self-implementation results with open-source code.}
 \label{com_cifar}
\vskip -0.1in
\end{table*}

\section{Appendix C Additional Experiments}
\label{appendix_c}

\subsection{Ablation experiments using MS-ResNet and Spike-driven Transformer architectures}
\label{appendix_c1}
Ablation studies of the TRR method with MS-ResNet18 and Spike-driven Transformer architectures are shown in Table~\ref{ablation:resnet} and Table~\ref{ablation:sdt}. It can be seen that our TRR method improves the performance of the baseline on both architectures. Specifically, using MS-ResNet18 on the CIFAR10-DVS, TRR improved the accuracy of the baseline by 8.20\%, which is a significant improvement.

It is worth noting that when using the Spike-driven Transformer architecture on CIFAR10/100, the use of feature hybridization alone caused a degradation in model performance. We speculate that this is due to overly strong perturbations, just as overly strong regularization can lead to model underfitting, which may require careful tuning of the balance coefficient $\alpha$. Fortunately, when incorporating feature hybridization and 
temporal reversal, TRR still contributes positively to the performance of the model and performs better than temporal reversal alone. This is because the consistency loss in temporal reversal enhances the representation of the model, moderating the negative impact of feature hybridization and turning it into a positive facilitation effect.

\begin{figure*}[!t]
\centering
\includegraphics[width=0.98\textwidth]{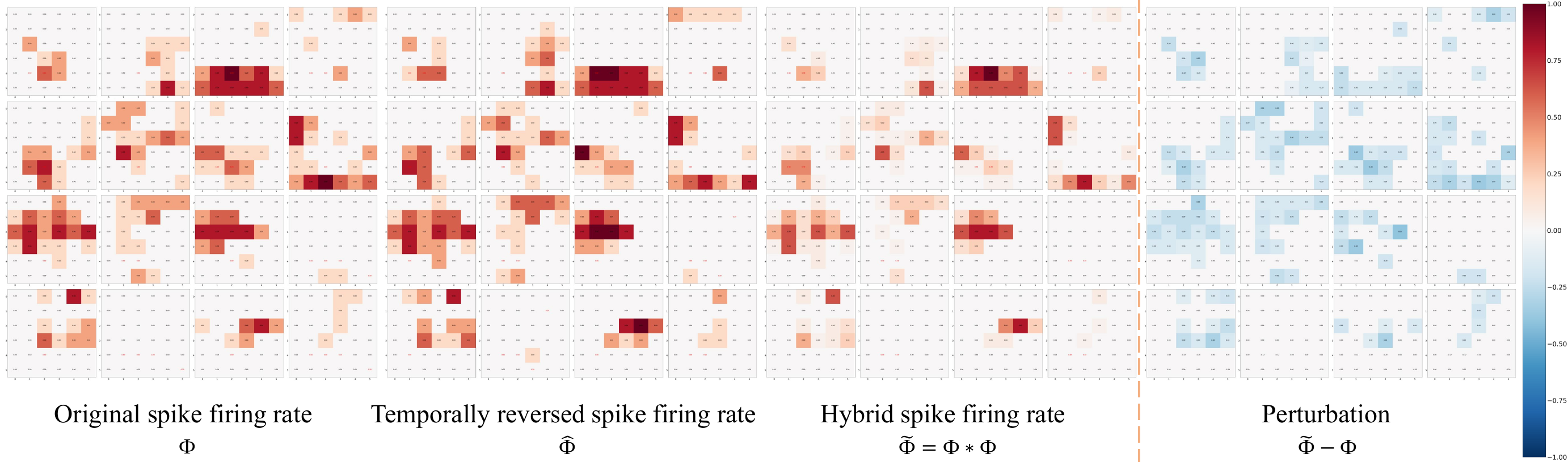}
\vskip -0.1in
\caption{Visualization of original, temporally reversed, and hybride spike firing rates and perturbations after 1 epochs of TRR training. Shown here are the first 16 channels of the penultimate layer of the VGG-9 network on the CIFAR10-DVS, where the input is the example in Fig.~\ref{visfull}. The results show that the ``star" operation hybridization caused a significant negative perturbation (blue area in the rightmost subfigure).}
\label{visstar_epoch1}
\end{figure*}

\begin{figure*}[!t]
\centering
\includegraphics[width=0.98\textwidth]{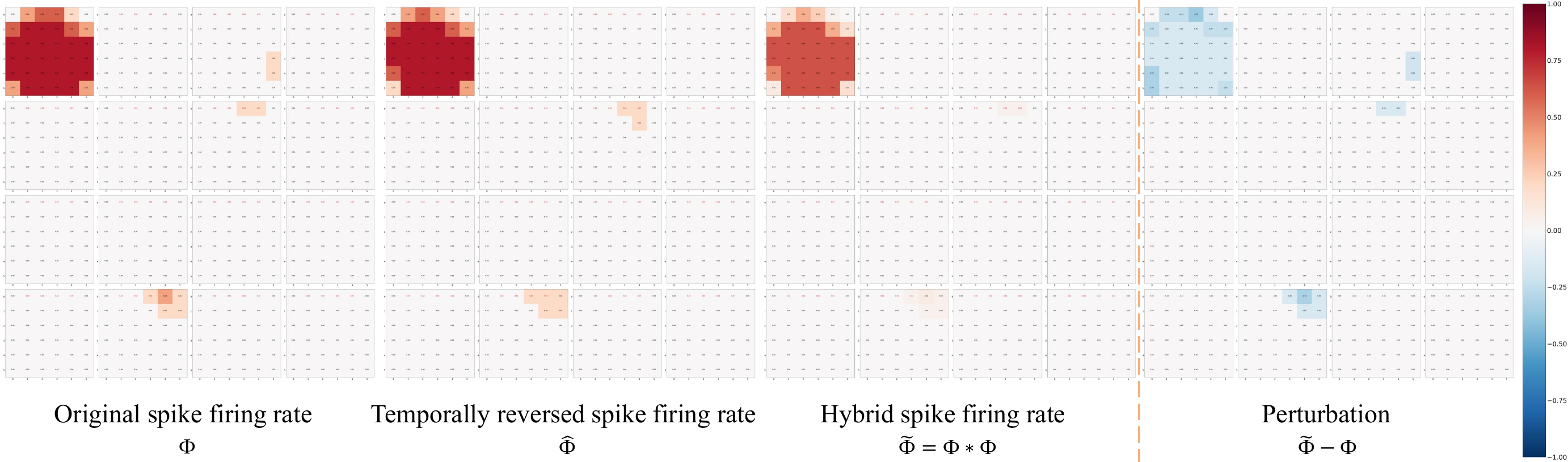}
\vskip -0.1in
\caption{Visualization of original, temporally reversed, and hybride spike firing rates and perturbations after 100 epochs of TRR training. Shown here are the first 16 channels of the penultimate layer of the VGG-9 network on the CIFAR10-DVS, where the input is the example in Fig.~\ref{visfull}. The ``star" operation hybridization induced fewer negative perturbations than at the beginning of training, indicating that the model learned perturbation-insensitive generalized representations.}
\label{visstar}
\end{figure*}

\begin{table}[t]
 \centering
 \begin{tabular}{cccc}
  \toprule
 Dataset & Baseline & Spike star & Firing rate star\\
  \midrule
  CIFAR10 & 93.67 & 93.94 & 94.02\\
  CIFAR10-DVS & 73.39 & 74.93 & 75.33\\
  \bottomrule
 \end{tabular}
 \vskip -0.1in
 \caption{Comparative results of spike star and spike firing rate star (\%). Too sparse spikes result in a weaker performance of the direct spike ``star" than the spike firing rate ``star".}
 \label{spikestar}
\end{table}

\subsection{Comparison of Spike Star and Spike Firing Rate Star}
\label{appendix_c2}
We investigated the performance of direct spike hybridization using ``star operation" on CIFAR10 and CIFAR10-DVS with VGG-9, and the results are shown in Table~\ref{spikestar}. The results show that direct ``star" hybridization of binary spike features can also improve model performance, but the sparse spikes cause the hybridization results to not be well regularized, and thus the performance is weaker than spike firing rate hybridization.

\subsection{Comparative Results on CIFAR}
\label{appendix_cifar}

The comparative results of TRR and other methods on CIFAR10 and CIFAR100 are shown in Table~\ref{com_cifar}. Our method achieved 96.71\% and 81.65\% accuracy on CIFAR10 and CIFAR100, respectively, using QKFormer, surpassing other methods. Even with ResNet and VGG architectures, TRR delivers competitive performance.

\section{Appendix D Additional Visualizations}
\label{app_vis}

\begin{figure*}[!t]
\centering
\includegraphics[width=0.98\textwidth]{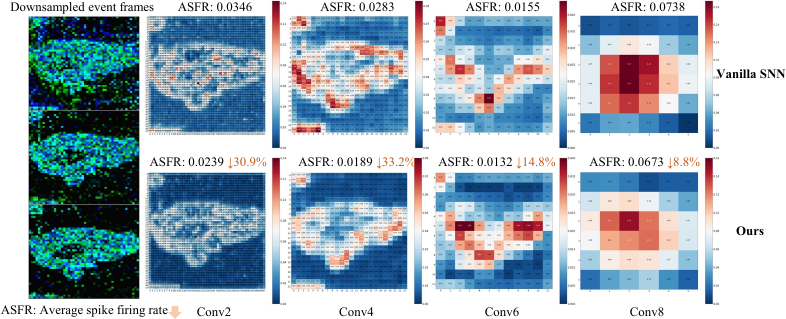}
\vskip -0.1in
\caption{Visualization of ASFR on CIFAR10-DVS with VGG-9. Our method simultaneously achieves higher performance and lower ASFR, which reduces energy consumption during deployment. The ASFR was reduced by 8.8\% to 33.2\% for the four stages.}
\label{visfull}
\end{figure*}

\begin{figure*}[!t]
\centering
\includegraphics[width=0.98\textwidth]{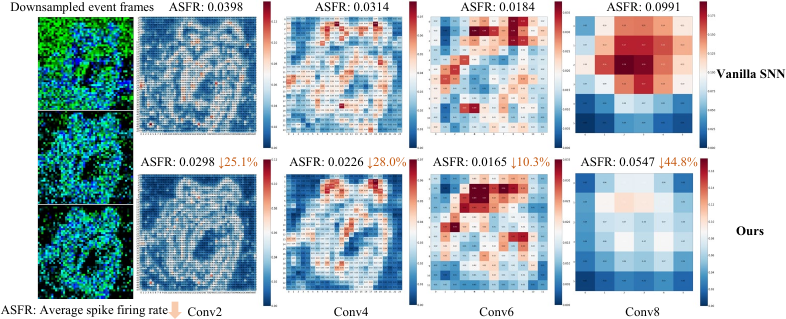}
\vskip -0.1in
\caption{Visualization of ASFR on CIFAR10-DVS with VGG-9. Compared to the baseline, our method reduced the ASFR by 10.3\% to 44.8\%.}
\label{visfull2}
\end{figure*}

\subsection{Visualization of Spike Firing Rate Perturbation}
\label{appendix_d1}

Using the first 16 channels of the penultimate layer of VGG-9 on CIFAR10-DVS, the original, temporally reversed, and hybrid spike firing rate, as well as the visualization of the perturbation results, are shown in Fig.~\ref{visstar_epoch1} and Fig.~\ref{visstar}. The model in Fig.~\ref{visstar_epoch1} was trained for only one epoch, and the model in Fig.~\ref{visstar} is a well-trained model. We obtain perturbation information by subtracting the original spike firing rate $\Phi$ from the hybrid spike firing rate $\tilde{\Phi}$. It can be seen that the perturbations caused by the ``star operation" have a significant influence, especially for the model trained with only one epoch. As training continues, the model extracts generalization features that are less sensitive to perturbations, and this hybridization produces smaller and smaller perturbations. In addition, since the hybridization also results from the temporally reversed spike firing rate, there is an inherent effect of temproal perturbation. Considering these two points, this hybridization can be considered as a spatio-temporal regularization that facilitates the generalizability of the SNN.

\subsection{Visualization of Average Spike Firing Rate}
\label{appendix_d2}
The ASFR of VGG-9 at four stages on CIFAR10-DVS is shown in Fig.~\ref{visfull} and Fig.~\ref{visfull2}, where our TRR significantly reduced the ASFR while achieving better performance than the baseline. Specifically, for these four stages, TRR reduced ASFR by 8.8\% to 44.8\% compared to the baseline. When the SNN is deployed on a neuromorphic chip, the inference energy consumption of the SNN depends entirely on the number of spikes, i.e., the ASFR is positively correlated with the inference energy consumption. Therefore, our method reduces the energy consumption of SNNs and is more favorable for deployment on resource-constrained edge devices.

\begin{figure*}[!t]
\centering
\includegraphics[width=0.98\textwidth]{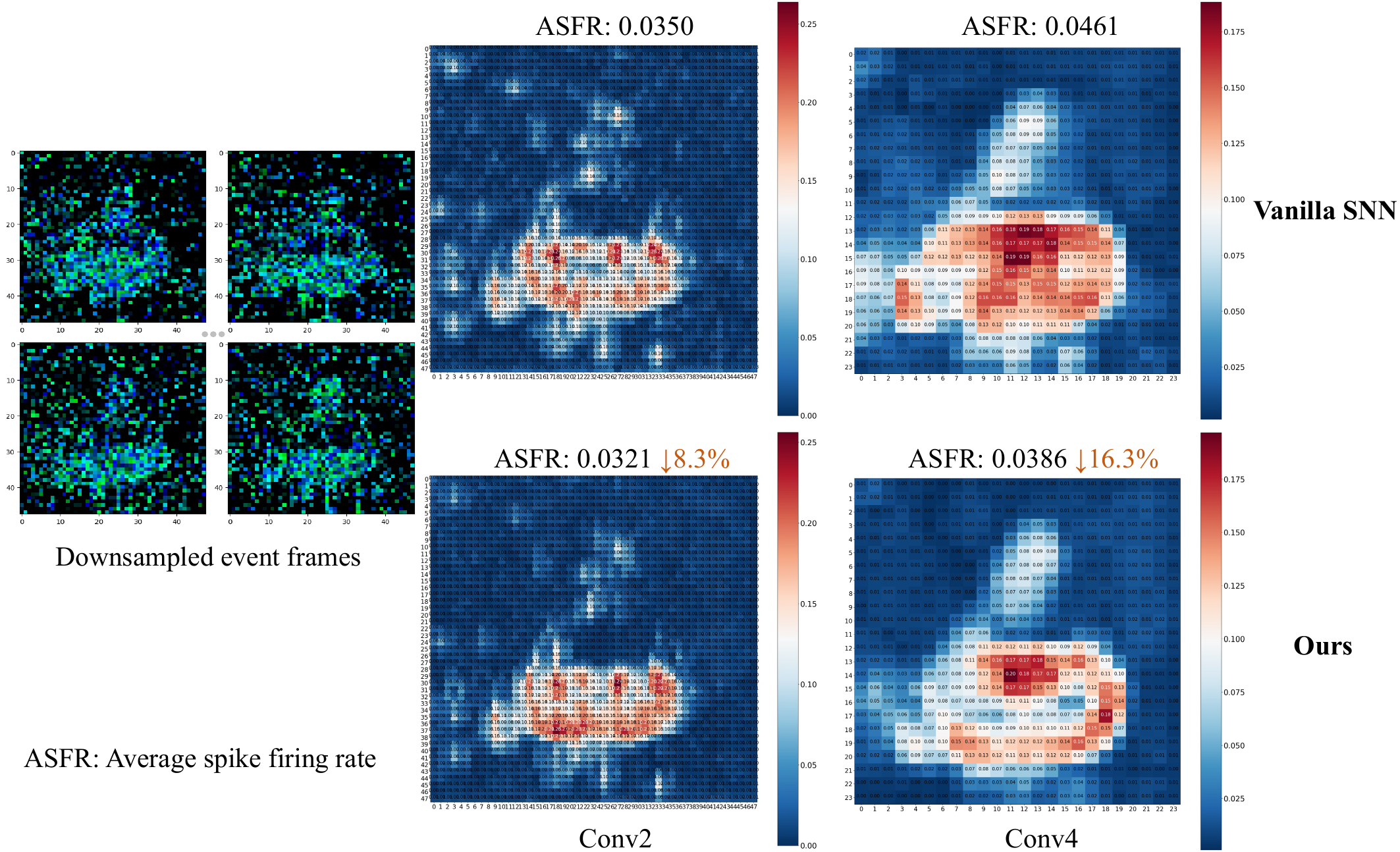}
\vskip -0.1in
\caption{Visualization of ASFR on DVS-Gesture with VGG-9. Compared to the baseline, our method reduced the ASFR by 8.3\% to 16.3\%.}
\label{visgesture}
\end{figure*}

Additionally, the ASFR visualization results on DVS-Gesure are shown in Fig.~\ref{visgesture}. Similar to on CIFAR10-DVS, our TRR has a lower ASFR than the baseline model, further supporting the high-performance, low-energy advantage of our TRR method.

\end{document}